\documentclass{article}

     \usepackage[final,nonatbib]{neurips_2019}

\usepackage[utf8]{inputenc} 
\usepackage[T1]{fontenc}    
\usepackage{hyperref}       
\usepackage{url}            
\usepackage{booktabs}       
\usepackage{amsfonts}       
\usepackage{nicefrac}       
\usepackage{microtype}      
\usepackage{graphicx}
\usepackage{amsmath}

\title{Human-centric Metric for \\ Accelerating Pathology Reports Annotation}

\author{%
  Ruibin Ma\thanks{Work done at Google Health}\\
  UNC Chapel Hill
  \vspace{-3em}
   \And
   Po-Hsuan Cameron Chen\\
   Google Health
   \And
   Gang Li\\
   Google Health
   \And
   Wei-Hung Weng$^*$\\
   MIT\\
   \And
   Angela Lin\\
   Google Health
   \And
   Krishna Gadepalli\\
   Google Health
   \And
   Yuannan Cai\\
  Google Health
}

\begin{document}
\maketitle

\vspace{-8pt}
\begin{abstract}
\vspace{-8pt}

Pathology reports contain useful information such as the main involved organ, diagnosis, etc. These information can be identified from the free text reports and used for large-scale statistical analysis or serve as annotation for other modalities such as pathology slides images. However, manual classification for a huge number of reports on multiple tasks is labor-intensive. 
In this paper, we have developed an automatic text classifier based on BERT and we propose a human-centric metric to evaluate the model. According to the model confidence, we identify low-confidence cases that require further expert annotation and high-confidence cases that are automatically classified. We report the percentage of low-confidence cases and the performance of automatically classified cases. On the high-confidence cases, the model achieves classification accuracy comparable to pathologists. This leads a potential of reducing 80\% to 98\% of the manual annotation workload. 
\end{abstract}

\vspace{-8pt}
\section{Introduction}
\vspace{-8pt}

Pathology is the gold standard for medical diagnosis, especially for diagnosing malignancy. Recent advances in machine learning (ML) for pathology have shown  improved diagnostic accuracy and efficiency for pathologists \cite{jama,naturemedarm}, and even discover unknown signals that are useful for clinical decision making \cite{naturemedmsi}. These studies rely on well-curated datasets. In this study, we explore using machine learning approach to assist with pathology report annotation. 

Among all data in pathology, free text reports contain rich information including the examined organ, histological findings, diagnosis, and other tissue-level or molecular-level evidence.
Furthermore, they comprehend pathologists' domain knowledge since the content of reports are curated from complicated information from both gigapixel-level slide images and the clinical narratives. Thus, the well-organized content from pathology reports are essential for not only developing a system for pathology slide archiving and cohort selection, but also associating to the corresponding slide images and serving as features for ML tasks in pathology \cite{naturemedarm,breastnodalmetastasis,cancermetastasis}. 

However, curating and annotating the free text reports often require manual review to understand the case composition, which is time-consuming,  labor-extensive, and non-scalable. 
%
%
%
Since the reports consist of plentiful domain-specific terminologies, even the well-trained clinicians in other medical domains might have difficulties in understanding them \cite{weng2019unsupervised}. 
Such limitations yield a greater barrier to accelerate the annotation and utilize the informative reports. 
To overcome the limitations, we re-framed the annotation task into an text classification problem based on ML.

Automatic text classification is an important ML task in which we have seen many breakthroughs driven by natural language processing (NLP). In the medical domain, NLP has been useful for classifying clinical narratives into medical subdomains \cite{subdomain}, which is also metadata information of clinical notes. NLP can speed up the curation of oncologic outcomes from radiology reports \cite{radiologyreport}.

Researchers usually adopt recurrent or convolutional neural networks (RNN, CNN) to learn language representation due to their capability of capturing the patterns in sequential information \cite{wordpiece,kim2014convolutional}. Recently, Transformer \cite{transformer} achieved better performance, especially on longer sequences, using a fully attention-based mechanism. Bidirectional Encoder Representations from Transformers (BERT) \cite{bert} learned word or document representation by self-supervised training and outperformed other approaches across various NLP tasks including automated text classification \cite{glue}. 
However, the commonly-used metrics used in text classification, such as accuracy, precision, recall, F1 score and area under the receiver operating characteristic curve (AUC), do not provide an insight of model uncertainty and confidence from the user perspective. These metrics do not assess whether a case requires further human inspection.

In this paper we adopt the BERT encoder for pathology report classification, and we designed a new human-centric metric that measures the potential of expert annotation workload reduction. The proposed model automatically identifies low-confidence cases that require expert review. This is different from most of the prior studies that focused on ML-oriented metrics such as the AUC. We have conducted experiments on five text classification tasks and show that the model can reduce human workload by 80\% to 98\% while keeping the same accuracy level as the human expert on those high-confidence cases.

\vspace{-4pt}

\begin{figure}[!t]
    \vspace{-1.3em}
    \centering
    \includegraphics[width=0.9\textwidth]{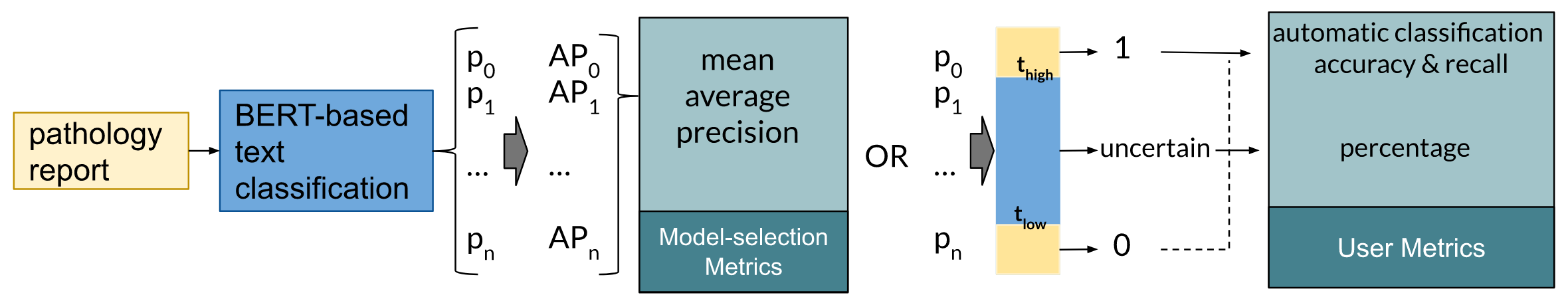}
    \caption{Study design of the proposed pipeline. We used the estimated case probability to compute both mean average precision for model selection and the human-centric metric for measuring model performance with potential of workload reduction.}
    \label{fig:flowchart} 
    \vspace{-1.3em}
\end{figure}

\section{Methods}

\vspace{-8pt}
\label{section:methods}
\subsection{Problem Formulation}

\vspace{-8pt}
We formulate the annotation problem as a multilabel classification problem with free text as input and a set of labels as output. The output can be an empty set if none of the labels are determined as positive. For example, if we are interested in the organs involved in a case, the input is the text of a pathology report while the outputs are the names of the organs. There can be more than one organ examined in a case, such as lymph node and breast for a case of breast cancer. 

\vspace{-8pt}
\subsection{Metrics}

\vspace{-8pt}
In this study, we consider two metrics: a model-selection metric for model development and a user metric for evaluating the potential utility for automatic case classification (Figure \ref{fig:flowchart}).

\paragraph{Model-selection metric:} 
We use the mean average precision (mAP) to select the final model for each task. It computes the mean of the average precision value for recall from 1 to 0 of each label \cite{precisionrecall}. Formally, suppose the task has a target label set $\{y_1, ... , y_n\}$, where $n$ is the number of labels. 
For each case, the model generates a score $p_i \in [0, 1]$ for label $y_i$. 
Based on these scores $p_i$, an average precision ($AP_i$) score can be calculated by taking the area under the precision-recall (PR) curve. The mean average precision is computed by $mAP=\frac{1}{n}\sum_{i=1}^n AP_i$. In this study, we use the composite trapezoidal method with intermediate PR points to compute average precision scores \cite{carefulinterpolation}.

\paragraph{User metric:} 
We designed a user metric to measure the potential utility of the model for accelerating expert annotation. In practice, we classify each pathology report into either the low model confidence group, which requires further expert inspection, or the high model confidence group that can bypass manual annotation. We report the percentage of the low-confidence cases to estimate the remaining human workload. We expect the classifier to perform well on the cases in the high-confidence group. For the high-confidence group, we report the accuracy and recall for the model performance. 

In detail, given the predicted score $p_i$, we set two thresholds, $t_{\mathrm{low}}$ and $t_{\mathrm{high}}$, for splitting the low and high-confidence groups. A case with no score between $t_{\mathrm{high}}$ and $t_{\mathrm{low}}$ is deemed a high-confidence case. On high-confidence cases we compute the so-called ``subset accuracy'', where all labels of a sample should be classified correctly for it to be considered correct. We also compute the mean of the recalls of all labels. This is more strict than a sample-wise recall when the dataset is unbalanced, which is one of the challenges in our tasks. If a case has any label whose score is between $t_{\mathrm{low}}$ and $t_{\mathrm{high}}$, that case will be considered as a low-confidence case. It will be passed to an expert for annotation. (Appendix \ref{appendix:tuningthresholds} for the tuning strategy of the two thresholds.) 

\subsection{Model for Text Encoding and Classification}

The model includes a text encoder and a classifier. The encoder is a BERT-based model with 12 Transformer blocks, 768 hidden states and 12 self-attention heads \cite{bert}. It converts an free text input sentence into a 2048-D embedding vector. The classifier contains a fully-connected layer which converts the embedding into a logits vector of length equal to the number of target classes and a sigmoid layer which turns the logits into a prediction score vector from 0 to 1.

We use a weighted categorical cross-entropy loss to fine-tune the encoder and train the classifier: $\mathcal{L}= \sum_{k=1}^{m}\sum_{i=1}^n w_i(\hat{y}_{i}^k\log p_{i}^k + (1-\hat{y}_i^k)\log(1- p_{i}^k))$, where $m$ is the number of samples; $n$ is the number of labels; $p_i^k$ and $\hat{y}_i^k$ are the prediction score and the ground truth label (0 or 1) of the $i$-th label of sample $k$. To calibrate for the imbalanced dataset during training, we design $w_i$ to be a label-wise weight pre-computed on the whole training set. It is either 1.0 for uniform weightings or a value inversely proportional to the number of positive samples of label $i$ for ``balanced weighting''. 

\vspace{-8pt}
\section{Experiments}

\vspace{-8pt}
\paragraph{Dataset}
We collected 290,438 pathology reports between 2005 and 2015 from a tertiary teaching hospital in the United States. Among them, 99,470 de-identified reports between 2005 and 2010 were manually annotated on the five tasks by board-certified pathologists. The five tasks are main organ (15 classes), disease type (3 classes), cancer reason (6 classes), primary cancer site (15 classes) and metastatic disease (3 classes). The exact size of the dataset may differ depending on the task. For example, for the ``cancer reason'' prediction, we only use 8,870 cases who have cancer. The detail of the task labels and the numbers of samples are in Appendix \ref{appendix:taskdetails}.

We split the dataset of each task into training (65\%), validation (15\%) and testing (20\%) sets. We selected the final model for each task by taking the model with the best model-selection metric on the validation set. These final models were further evaluated by both model-selection and user metric on the testing set. 


\vspace{-8pt}
\paragraph{Different pretrained models:}
Due to the difference of word and semantic distribution between general English and pathology, we investigated three configurations for pretraining BERT encoder:

1) \textit{base}: the BERT encoder pretrained by English corpora with an English vocabulary of size 30,522 \cite{bert}.

2) \textit{base+pathology}: On top of \textit{base}, we further pretrained the BERT encoder using our pathology reports with the same English vocabulary. 

3) \textit{pathology}: We trained a randomly initialized BERT encoder from scratch using our pathology corpora and a designated pathology vocabulary with the size of 9,755. We used the same training data as \textit{base+pathology} with a different tokenizer due to a different vocabulary.
	
The above-mentioned vocabularies were learned from general English or pathology corpora by the WordPiece tokenization \cite{wordpiece}. The pathology corpus was built from all of the pathology reports (including the unlabelled reports) that were not used in the validation and test sets. Table \ref{table:initialmodel} compares $mAP$ across the three different initial BERT encoders on the validation set. The encoders pretrained on pathology corpus yielded better results on all tasks. Pretraining with domain-specific corpus on top of the general BERT encoder (\textit{base+pathology}) outperforms the encoder trained from scratch on 4/5 tasks.

\small
\begin{table}[t!]
\centering
\resizebox{\textwidth}{!}{
 \begin{tabular}{crrrrr} 
 \toprule
 Configuration & Main organ & Disease type & Cancer reason & Primary cancer site & Metastatic \\ [0.5ex] 
 \midrule
 \textit{base} & 0.9158 & 0.9221 & 0.9118 & 0.8762 & 0.8459 \\ 
 \textit{base+pathology} & \textbf{0.9267} & \textbf{0.9240} & \textbf{0.9668} & 0.9043 & \textbf{0.9097} \\
 \textit{pathology} & 0.9197 & 0.9182 & 0.9337 & \textbf{0.9130} & 0.8802 \\
 \bottomrule
 \end{tabular}
}
\vspace{1pt}
\caption{Comparison of mean average precision ($mAP$) of each classification task with different configurations for pretraining BERT encoder.}
\label{table:initialmodel}
\vspace{-20pt}
\end{table}
\normalsize

\vspace{-8pt}
\paragraph{Compare mean average precision with baselines:} We compare our BERT-based method with a support vector machine (SVM) and logistic regression (LR) in an ``one-versus-rest'' approach \cite{bishopbook,scikit-learn}. The baselines treat a multilabel classification as a set of independent binary classifications and combine the positive labels as the result. We also compare the model performance with and without the balanced-weighting strategy (see Section \ref{section:methods}). The input texts were converted to term-frequency inverse-document-frequency (TF-IDF) features using 1,2,3-grams as the inputs of SVM and LR.

\begin{figure}[t]
    \vspace{-2em}
    \centering
    \includegraphics[width=0.95\textwidth,height=3.5cm]{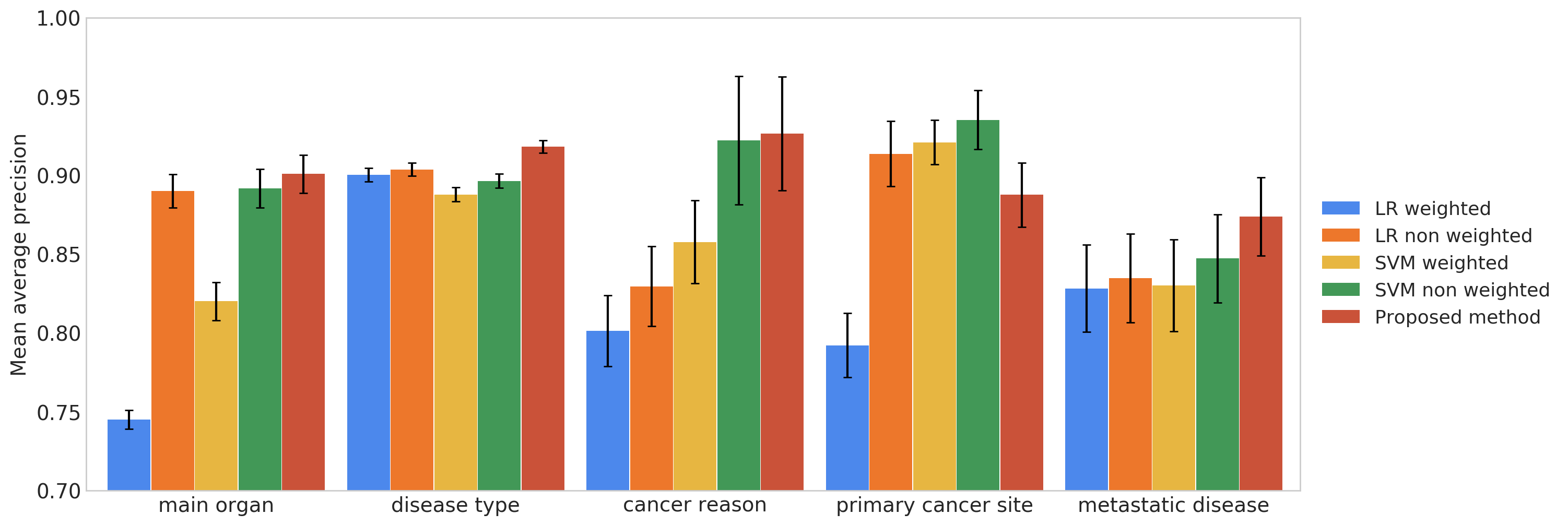}
    \vspace{-8pt}
    \caption{Model performance using different machine learning methods across five tasks.}
    \label{fig:mAP} 
    \vspace{-2pt}
\end{figure}

Figure \ref{fig:mAP} shows comparisons between the proposed method and the baselines across five tasks on the test set. The proposed model outperforms the other models in 4/5 tasks. The only task without improvement is the primary cancer site prediction. One reason could be that the data distribution of validation and test sets are not similar on this task and the model overfit to the validation set. This inconsistency may be caused by fewer data points (only 8k samples for this task because it is only applicable to cancer positive cases) and a large but very unbalanced target label set (15 classes) (Appendix Table \ref{table:taskdetails}). The number of the positive cases for some labels can be very limited.

\vspace{-8pt}
\paragraph{Evaluation of user metric:}
Table \ref{table:usermetrics} shows the evaluation of the user metric across the five tasks. We mainly show the percentage of the low-confidence cases and the accuracy of the high-confidence (automatically classified) cases. We estimated the human accuracy by calculating the consistency among three pathologists on a sampled set of reports. The consistency is defined as follows: $\mathrm{consistency}=\frac{1}{N}\sum_{i=1}^N c_i$, where $N$ is the number of reports and $c_i$ is the consistency score of report $i$. $c_i=\frac{1}{3}$ if the three annotations are inconsistent; $c_i=\frac{2}{3}$ if two of the three are consistent; $c_i=1$ if all three are consistent. For comparison, we also show the accuracy of automatic classification on the full test set. 

In Table \ref{table:usermetrics}, we demonstrate that the proposed model has comparable performance in the high-confidence group as board-certified pathologists (better on 3/5 tasks). The performance became inferior when we tested on the full test set. This indicates that our strategy of filtering the low-confidence cases is effective. Human labor can be reduced by 80\% to 98\% depending on tasks.

\small
\begin{table}[t]
\begin{center}
 \begin{tabular}{crrrrr} 
 \toprule
 Task & \shortstack[c]{\\Uncertain \\percentage} & \shortstack[c]{\\Automatic \\classified\\ average\\ recall} & \shortstack[c]{\\Automatic \\classified\\ accuracy} & \shortstack[c]{\\Human\\ accuracy \\(consistency)} & \shortstack[c]{\\Accuracy \\on full \\test set} \\ 
 \midrule
 \textit{main organ} & 20.24\% & 84.84\% & \textbf{95.70\%} & 95.31\% & 91.15\% \\ 
 \textit{disease type} & 10.99\% & 85.29\% & \textbf{96.20\%} & 95.42\% & 90.25\% \\ 
 \textit{cancer reason} & 1.93\% & 79.66\% & 99.18\% & \textbf{99.24\%} & 98.47\% \\ 
 \textit{primary cancer site} & 1.99\% & 83.72\% & \textbf{97.51\%} & 97.39\% & 96.70\% \\ 
 \textit{metastatic disease} & 1.82\% & 83.44\% & 98.61\% & \textbf{99.21\%} & 98.01\% \\ 
 \bottomrule
 \end{tabular}
\end{center}
\caption{Comparison between human expert and the performance of classification model in the high-confidence group identified by the designed user metric.}
\label{table:usermetrics}
\vspace{-16pt}
\end{table}
\normalsize


\vspace{-8pt}
\paragraph{Conclusion}
In this study, we designed a practical and intuitive human-centric metric to accelerate the process of pathology report annotation.
The metric helps identify the high-confidence samples that are suitable for automated classification, and low-confidence samples require further expert annotation.
We showed that the proposed model reduced expert effort by 80\%- 98\% while achieving comparable (slightly better) performance of pathologists on the automated text classification tasks in the high-confidence sample group identified by the user metric.

There are several limitations in this study. The dataset was highly unbalanced (Appendix B), which led to the difficulty in learning of classes with few samples. This was reflected in the low average recall in Table \ref{table:usermetrics}. Some potential solutions include data augmentation and hard negative mining. Data augmentation creates synthetic data by approaches like random sentence modification on the existing dataset. Hard negative mining oversamples the difficult cases in the training time. Secondly, the study used a dataset from a single tertiary teaching hospital across more than ten years. It is unclear how well the model can be generalized to different institutions. We plan to conduct external validation to make the proposed method more robust.
\newpage



\small
\bibliographystyle{splncs04}
\bibliography{reference}

\newpage
\appendix
\setcounter{table}{0}
\renewcommand{\thetable}{A\arabic{table}}

\section{Tuning thresholds}
\label{appendix:tuningthresholds}
We let $t_{\mathrm{high}} = 1 - t_{\mathrm{low}}$ and search for $t_{\mathrm{low}}$ within $[0.01, 0.05, 0.1, 0.2, 0.3, 0.4, 0.45]$. We choose the value with highest $\mathrm{automatic\,accuracy} \times (1 - \mathrm{uncertain\, percentage})$.

\section{Task details}
\label{appendix:taskdetails}
The task details such as number of cases of each dataset and number of positive cases of each label are shown in Table \ref{table:taskdetails}.

\begin{table}[h]
\begin{center}
\small
\scalebox{0.9}{
 \begin{tabular}[\textwidth]{| l | l | l |} 
 \hline
 task & number of cases of each dataset & (number of positive cases) labels/classes\\ [0.5ex] 
 \hline
 main organ & \shortstack[l]{\\ all: 89001 \\ train: 57850 \\ val: 13349 \\ test: 17802 \\ human: 5464} &
 \shortstack[l]{\\
 (3097) Breast \\
 (297) Lung or bronchus \\
 (1751) Prostate gland \\
 (7541) Colorectal (colon and rectum) \\
 (26001) Skin \\
 (1470) Lymph node \\
 (379) Kidney \\
 (18032) Uterus (corpus and cervix) \\
 (860) Ovary \\
 (505) Testis \\
 (792) Liver \\
 (36) Pancreas \\
 (494) Thyroid gland \\
 (7083) Head and neck \\
 (6264) Upper GI 
 } \\
 \hline
 disease type & \shortstack[l]{\\ all: 89001 \\ train: 57849 \\ val: 13348 \\ test: 17804 \\ human: 5464} &
 \shortstack[l]{\\
 (74973) Non-Cancer \\
 (9087) Pre-malignant \\
 (8831) Cancer
 } \\
 \hline
 cancer reason & \shortstack[l]{\\ all: 8870 \\ train: 5765 \\ val: 1329 \\ test: 1776 \\ human: 504} &
 \shortstack[l]{\\
(7999) Carcinoma \\
(299) Melanoma \\
(155) Soft Tissue/Sarcoma \\
(85) Germ Cell Tumors \\
(33) Blastomas and other primitive tumors \\
(200) Lymphoma and hematologic cancers
 } \\
 \hline
 primary cancer site & \shortstack[l]{\\ all: 8870 \\ train: 5764 \\ val: 1330 \\ test: 1776 \\ human: 504} &
 \shortstack[l]{\\
 (857) Breast \\
 (106) Lung or bronchus \\
 (693) Prostate gland \\
 (276) Colorectal (colon and rectum) \\
 (5097) Skin \\
 (132) Lymph node \\
 (126) Kidney \\
 (380) Uterus (corpus and cervix) \\
 (38) Ovary \\
 (78) Testis \\
 (16) Liver \\
 (12) Pancreas \\
 (208) Thyroid gland \\
 (266) Head and neck \\
 (88) Upper GI 
 } \\
 \hline
 metastatic disease & \shortstack[l]{\\ all: 8870 \\ train: 5765 \\ val: 1330 \\ test: 1775 \\ human: 504} &
 \shortstack[l]{\\
(8245) No\\
(418) Yes: In lymph nodes\\
(207) Yes: In non-lymph node tissue(s)
 } \\
 \hline
 \end{tabular}
 }
\end{center}
\caption{Task details: "human" refers to the number of cases used to estimate human consistency. The last three tasks are cancer-related tasks so their number of cases are much fewer than the first two tasks. The number of "train", "val" and "test" can be slightly different between the first two tasks and between the last three tasks because of stratification.}
\label{table:taskdetails}
\end{table}







\end{document}